\documentclass[a4paper]{article}

\usepackage{fullpage} 
\usepackage{parskip} 
\usepackage{tikz} 
\usepackage{amsmath}
\usepackage{hyperref}
\usepackage{subfig}

\usepackage{graphicx}             
\usepackage{moreverb}             

\usepackage{graphicx}
\usepackage{amsmath,amssymb} 
\usepackage{color}
\usepackage{blindtext}
\usepackage{tabto, hyperref}
\usepackage{subfig, array}
\usepackage{blindtext, amsmath}
\usepackage{graphicx, tabto, hyperref}
\usepackage{subfig}
\usepackage{dblfloatfix}
\usepackage{graphics}
\usepackage{fontenc}
\usepackage{txfonts}
\usepackage{pifont}
\usepackage{color}
\usepackage{hyperref}
\usepackage{booktabs}
\usepackage{multirow}
\usepackage{float}
\usepackage{siunitx, array}
\usepackage{authblk}

\usepackage{float}
\usepackage{siunitx}
\usepackage[linesnumbered,ruled]{algorithm2e}

\usepackage{array}
\newcolumntype{P}[1]{>{\centering\arraybackslash}p{#1}}

\title{Contrastive Fairness in Machine Learning}

\author[1]{Tapabrata Chakraborti}
\author[1]{Arijit Patra}
\author[1]{Alison Noble}

\affil[1]{Dept. of Engineering Science, University of Oxford, UK}

\begin{document}

\maketitle

\begin{abstract}


Was it fair that Harry was hired but not Barry? Was it fair that Pam was fired instead of Sam? How can one ensure fairness when an intelligent algorithm takes these decisions instead of a human? How can one ensure that the decisions were taken based on merit and not on protected attributes like race or sex? These are the questions that must be answered now that many decisions in real life can be made through machine learning. However research in fairness of algorithms has focused on the counterfactual questions ``what if?" or ``why?", whereas in real life most subjective questions of consequence are contrastive: ``why this but not that?". We introduce concepts and mathematical tools using causal inference to address contrastive fairness in algorithmic decision-making with illustrative examples.

\end{abstract}

\section{Introduction}

\noindent Machine learning based decision systems have achieved near human performance in many tasks in recent times. But as these algorithms have grown more powerful, they have become more complex (with numerous parameters) and hence more opaque (the decision making process is not easily explainable) [1]. Machine learning, after all, is a data driven optimal function fitting exercise, thus it has mostly dealt with association, rather than causation [2]. Given the broad use of machine learning algorithms in the modern world, precautions to ensure the fairness of the decision making process of such algorithms is of great importance [3].

The algorithm may take decisions partly based on protected variables (race, gender, sexual orientation, etc) learned from historic data having inherent bias [4]. Then there is the possibility of such bias getting perpetuated with significant social consequence for such tasks like job recruitment, university admission, insurance/lending, preemptive criminal profiling, etc [5, 6] to name a few. Modern machine learning methods should avoid such unethical discriminatory practice [7]. After all, the efficacy of a decision making process should be based on both efficiency and ethics [8].

We present \emph{contrastive fairness}, a new direction in causal inference applied to algorithmic fairness. Earlier causal inferential methods in algorithmic fairness dealt with the "what if?" question [9]. We establish the theoretical and mathematical foundations to answer the contrastive question "why this and not that?". This is essential to defend the fairness of algorithmic decisions in tasks where a person or sub-group of people is chosen over another (job recruitment, university admission, etc). At its core, any question of fairness is a comparison, because equality is not absolute in society [10]. Some discrimination is part of the process itself (say employee recruitment), what must be ensured therefore, is that the discrimination is on fair grounds. Hence the question of why a certain person was chosen and not another, is of utmost pertinence. Contrastive questions and their explanation [11] have been around for quite some time but not within the purview of artificial intelligence and machine learning. Contrastive explanation in artificial intelligence has only recently been discussed in 2018 in [12] [21], but for the first time it is formally introduced to algorithmic fairness in this work. It is to be noted that the current paper is meant to lay theoretical and mathematical foundations of contrastive logic in the realm of algorithmic fairness with initial results.

\section{Background Concepts}

The present paper combines two distinct areas of research, those of algorithmic fairness and causal inference, both comparatively niche areas in machine learning. We provide in this section for the reader, a brief collection of underlying definitions and concepts related to both of these areas.


\subsection{Algorithmic Fairness}

We first define a few notations which are used throughout this paper. Let $Y$ be the expected outcome and $\hat Y$ be the predicted outcome. $X$ is the set of observable attributes and $U$ is the set of latent attributes of an individual. $A$ is the set of protected attributes of the individual on which the algorithm should not base its prediction on, in order to be fair. Of course, the intuitive but naive assumption in that case would be that an algorithm maybe considered to be fair if $\hat Y$ is only dependent on $X$ and not $A$. However, this amounts to ``fairness through unawareness'' as there may be attribute(s) in $X$ that are analogous to attribute(s) in $A$, though not explicitly the same. This makes it necessary to devise more strict rules to ensure algorithmic fairness.

Most earlier notions of algorithmic fairness were global, that is true for the population. The two most popular among these are Demographic Parity and Equality of Opportunity. Demographic Parity holds if $P(\hat Y \mid A = 0) = P(\hat Y \mid A = 1)$, that is, we get the same prediction, irrespective of the value to which the protected attributes are set at. Note that this does not take into account the expected outcome $Y$ which means it ensures equality of result over the population, instead of any calibration using expected outcomes in sub-populations. Equality of opportunity does exactly that, it only seeks to ensure a certain prediction if the expected outcome supports that prediction for the sub-population in question. Equality of opportunity holds if $P(\hat Y = 1 \mid A = 0, Y =1) = P(\hat Y = 1 \mid A = 1, Y =1)$. It has been shown that these two criteria can never simultaneously hold true [13]. 


This brings to light the need for individual level fairness criterion, besides the above population level ones. If individual $i$ and $j$ are similar, that is some distance metric $d(i,j)$ is less than a small threshold, then individual fairness holds if $\hat{Y}(X^{(i)}, A^{(i)}) \approx \hat{Y}(X^{(j)}, A^{(j)})$. Of course, this introduces the constraint that the metric $d(i,j)$ should be properly chosen, which requires some domain knowledge expertise.

\subsection{Causal Inference}
\label{subsec:cmc}

Structural Causal Models (SCM) [14] are the backbone of causal inference methods [15]. These consist of three major interacting elements: causal diagrams, structural equations, and counterfactual/intervention logic. These together make up the triple of sets $(U,X,F)$ which constitute the SCM.


\begin{enumerate}
    \item Causal graphical diagrams are basically directed acyclic graphs (DAG). The nodes of the diagram are the variables and the directed arrow between them specify the flow of causal relations between the variables. There are two types of variables: $U$ is the set of latent background variables and $X$ are observable variables.
    \item Structural equations are a set of functions $\{f_1, \dots, f_n\} \in F$ corresponding to the variables $\{X_1, \dots, X_n\} \in X$ such that $X_i = f_i(p_i, U_{p_i})$, $p_i \subseteq X \backslash \{X_i\}$ and $U_{p_i} \subseteq U$.
    \item For causal inferential analysis, counterfactual and interventional logic are carried out using a set of rules called \emph{do-calculus}.
\end{enumerate}



Since causal diagrams are essentially directed acyclic graphs, each observable variable $X_i$ will be connected to its parent variables $p_i$, where $X_i \subseteq X$ and $p_i \subseteq X \backslash {X_i}$. Thus we see above that the value of the observable variable $X_i$ depends on its parent variables as well as the latent variables $U$, through the function $f_i$.

\textbf{Intervention logic.} As seen above, the value of a measurable variable $X_i$ is given by $X_i = f_i(p_i, U_{p_i})$. Now if an external agent deliberately sets the value of $X_i = x$, then that is called an intervention (eg. randomised control trials). So assuming that we know the probability distribution $P(U)$ of latent variables $U$, we can perform an intervention on $Z$ variables belonging to $X$ (that is $Z \subseteq X$), and then compute the resulting probability distribution of the remaining variables in $X$ other than $Z$, that is $X \setminus Z$.


\textbf{Counterfactual logic.} This then also helps us to do counterfactual calculations, where we essentially compute $P(Y_{Z \leftarrow z}(U)\ |\ W \!=\! w)$. Here $Y$ are those variables belonging to the set of observable variables $X$, that we want to measure the effect of, so essentially output variables. $Z$ are those variables that we have intervened on by setting to value $z$ and $W$ are all other variables with known probability distribution.

\subsection{Counterfactual Fairness}

Kusner \emph{et al.} [9] present the notion of counterfactual fairness. For a given problem of algorithmic fairness, let the causal model be given as usual by the set tuple $(U,V,F)$, where $V \equiv A \cup X$. $A$ are the protected variables and $U$ are the latent variables. $X$ are the observable variables other than $A$, so that they together make up the total set of observable variables $V$. $\hat Y$ is a fair predictor of the output variables $Y$ if

\begin{align}
     P(\hat Y_{A \leftarrow a\ }(U) = y\ |\ X = x, A = a)  = 
  P(\hat Y_{A \leftarrow a'}(U) = y\ |\ X = x, A = a)
\end{align}

This condition of counterfactual fairness should be fair for any $x, a, a'$ and for all $y$. The equation essentially enforces the condition that the probability distribution of $Y$ should not be affected if any of the protected variables are intervened on keeping other conditions the same [16].

\section{Contrastive Fairness}

In this Section we present the idea of contrastive fairness in detail which is the main contribution of this paper. First we present several contrastive fairness questions and then formulate them. We take initial cues from the work on contrastive explanation by Miller [12].

\subsection{Why do we need Contrastive Fairness? Why Counterfactual Fairness is not enough?}

Counterfactual fairness formalised the use of causal inference in ensuring fairness of machine learning algorithms. However, the criterion is population based, whereas many real life fairness questions compare how two individuals are treated, and whether the difference in decision for them was fair? Why was this decision taken for an individual and not some other decision? All these are contrastive cases of individual fairness, which requires some further considerations to be incorporated. We still use the same counterfactual logic but expand it to fit contrastive cases.

\subsection{Posing Contrastive Questions}

First we list the main contrastive causal questions from existing literature. Then we modify them into contrastive questions pertaining to algorithmic decision making. Lastly we modify them to form fairness criteria.

Van Bouwel and Weber [17] mention 3 kinds of \textbf{contrastive causal questions} that may be posed.

\begin{itemize}
\item \emph{P-contrast:} Why does object $X$ have property $P$, rather than property $Q$?
\item \emph{O-contrast:} Why does object $X$ have property $P$, but object $Y$ has property $Q$?
\item \emph{T-contrast:} Why does object $X$ have property $P$ at time $t$, but property $Q$ at time $t^{'}$?
\end{itemize}

This defines three types of contrast: within an object (P-contrast), between objects themselves (O-contrast), and within an object over time (T-contrast).

We modify these questions to ask \textbf{algorithmic decision questions} as follows:

\begin{itemize}
\item \emph{D-contrast:} Why does individual $I$ receive decision $D$, rather than decision $D'$? 
\item \emph{I-contrast:} Why does individual $I$ receive decision $D$ but individual $J$ receives decision $D'$? 
\item \emph{T-contrast:} Why does individual $I$ receive decision $D$ at time $t$, but decision $D'$ at time $t'$? 
\end{itemize}

Again, note the three types of contrast: for one individual (D-contrast), between two individuals (I-contrast), and for one individual over time (T-contrast).

Then we convert these into \textbf{contrastive fairness questions}, quite simply by rephrasing as follows:

\begin{itemize}
\item \emph{D-contrast:} Is it fair to make decision $D$ for individual $I$, instead of decision $D'$? 
\item \emph{I-contrast:} Is it fair to make decision $D$ for individual $I$, while make $D'$ for individual $J$? 
\item \emph{T-contrast:} Is it fair to make decision $D$ for individual $I$ at time $t$, but make $D'$ at time $t'$? 
\end{itemize}

These three questions framed here give the main groundwork of ensuring contrastive fairness of algorithmic decision making pertaining to an individual. Next we mathematically formulate these criteria.

\subsection{D-Contrast: Is it fair to make decision $D$ for individual $I$, instead of decision $D'$?}

This basically boils down to counterfactual fairness but for a particular individual. This is because if the decision making process is counterfactually fair for that individual for the entire decision space, then it should be fair when contrasting between any two decisions made for that individual. Thus the the decision making algorithm is fair for a particular individual $i$ if for any valid decision value $d$, the following holds.

\begin{align}
    P(\hat Y_{A_i \leftarrow a\ }(U_i) = d\ |\ X_i = x, A_i = a)  =   P(\hat Y_{A_i \leftarrow a'}(U_i) = d\ |\ X_i = x, A_i = a) 
\end{align}

Other symbols have same meaning as in eqn. 1. It should be noted that though the above equation ensures fairness of the decision making process, it does not comment on the fairness of the two competing decisions themselves. To show that the decision taken, say $d$, is better than an alternative decision say $d'$, the predicted probability score of the former should be greater than the latter as shown below.

\begin{align}
P(\hat Y(U_i) = d\ |\ X_i = x_i, A_i = a_i)  >
  P(\hat Y(U_i) = d^{'}\ |\ X_i = x_i, A_i = a_i) 
\end{align}

\subsection{I-Contrast: Is it fair to make decision $D$ for individual $I$, but $D'$ for individual $J$?}

When comparing decisions between two individuals however, we need to make further assumptions. Not only must the decision making processes be separately fair for both individuals, but also the difference in decision should be ``sensible", that is the probability values generated by the predictor should support that.

First we establish the fairness for the two individuals for the entire decision space as follows:

\begin{align}
   P(\hat Y_{A_i \leftarrow a_i\ }(U_i) = d\ |\ X_i = x_i, A_i = a_i)  =
  P(\hat Y_{A_i \leftarrow a'_i}(U_i) = d\ |\ X_i = x_i, A_i = a_i) 
\end{align}  

\begin{align}
 P(\hat Y_{A_j \leftarrow a_j\ }(U_j) = d\ |\ X_j = x_j, A_j = a_j)  = 
  P(\hat Y_{A_j \leftarrow a'_j}(U_j) = d\ |\ X_j = x_j, A_j = a_j) 
\end{align}

Next, even if the decision making process itself is fair, for the decision to ``make sense", for one individual the decision taken should have a higher probability score assigned by the predictor than the alternative decision, while the opposite should hold true for the other individual. This is presented mathematically as follows: 

\begin{align}
           P(\hat Y(U_i) = d\ |\ X_i = x_i, A_i = a_i)  > 
  P(\hat Y(U_i) = d^{'}\ |\ X_i = x_i, A_i = a_i)  
 \end{align}
  
\begin{align}
           P(\hat Y(U_j) = d'\ |\ X_j = x_j, A_j = a_j)  >  
  P(\hat Y(U_j) = d\ |\ X_j = x_j, A_j = a_j) 
\end{align}

Lastly, one must make sure that even if the protected variable values of the two individuals were to be same counterfactually, then also decision $D$ would have higher value than decision $D'$ for individual $I$ and decision $D'$ would have higher value than decision $D$ for individual $J$. This is present below.

\begin{align}
           P(\hat Y_{A_i \leftarrow a_j\ }(U_i) = d\ |\ X_i = x_i, A_i = a_i) 
  P(\hat Y_{A_i \leftarrow a_j\ }(U_i) = d^{'}\ |\ X_i = x_i, A_i = a_i)  
  \end{align}
  
\begin{align}  
           P(\hat Y_{A_j \leftarrow a_i\ }(U_j) = d'\ |\ X_j = x_j, A_j = a_j) 
  P(\hat Y_{A_j \leftarrow a_i\ }(U_j) = d\ |\ X_j = x_j, A_j = a_j) 
\end{align}

If these equations are satisfied then one can surmise that the contrast in decision made between these two individuals is fair.

\subsection{T-Contrast: Is it fair to make decision $D$ for individual $I$ at time $t$, but $D'$ at time $t'$?}

The main question being asked here is that if over time the decision made for an individual changes, is that change fair or not. To ensure this, one has to first verify that the process itself is fair for all valid decisions $d$ that can be made over all time points $t$. This is shown here:

\begin{align}
    P(\hat Y_{A_i \leftarrow a_i\ }(U_i) = d\ |\ X_i = x_i(t), A_i = a_i)  = 
  P(\hat Y_{A_i \leftarrow a'_i}(U_i) = d\ |\ X_i = x_i(t), A_i = a_i)
\end{align}

Now if the original decision was $d$ at time $t$, and became some other $d'$ at time $t'$, then it must also hold that at time $t$, decision $d$ had higher prediction than $d'$, whereas at time $t'$, the opposite is true. This is formulated below:

\begin{align}
           P(\hat Y(U_i) = d\ |\ X_i = x_i(t), A_i = a_i)  >  P(\hat Y(U_i) = d^{'}\ |\ X_i = x_i(t), A_i = a_i)
           \end{align}
           
  \begin{align}         
           P(\hat Y(U_i) = d'\ |\ X_i = x_i(t'), A_i = a_i)  > 
  P(\hat Y(U_i) = d\ |\ X_i = x_i(t'), A_i = a_i) 
\end{align}

\subsection{Illustrative thought example: Fairness of Employee Job Location}

Consider two employees $P$ and $Q$ having the same job duties and responsibilities in the same organisation. The organisation has office locations in London and other locations elsewhere in  the UK. Employees might have a preference of working in London, so if they are assigned to a different office location, the decision making process should be fair. This becomes even more significant for the organisation, if a contrastive allocation of location between two employees is challenged and needs to be defended, especially if the decision is taken by an algorithm based on employee background data and performance statistics. We discuss all three scenarios of contrastive fairness for this test case and consider what needs to be shown, in order to prove fairness.

Suppose employee $P$ is joining the organisation and in his/her application form had indicated that he/she would prefer to be located in the London headquarters, but is being located elsewhere due to more employee requirements in the satellite office, and this decision is being taken by a machine learning enabled "HR algorithm". This decision should not be based on such protected attributes like race, sex, religion, etc. This is an example of \emph{D-contrast} and to ensure fairness, the algorithm must satisfy equations 2 and 3. Also consider the case of the same employee $P$ being first located in London, and then at a later point of time being relocated to another office. In that case, the algorithm must satisfy the conditions of \emph{T-contrast} via equations 10, 11 and 12. Lastly consider employee $P$ being assigned a London office and employee $Q$ being assigned a satellite office and employee $Q$ counters this decision in the belief that the decision was biased by race. It is of great importance to the organisation to be able to justify the decision fairness of the "HR algorithm" and that can be achieved using the \emph{I-contrast} equations 4 to 9.

\subsection{Why contrastive decision instead of making same decision for both individuals?}

A core notion underpinning contrastive fairness is that one decision is more "desirable" than another and hence the need to justify the difference in decision between two individuals. In that case, a natural follow-up question that arises is why not make the "desirable" decision for both individuals, why at all go for the less desirable alternative decision? For example, revisiting the illustrative example in of job location, why not locate both employees $P$ and $Q$ in the London office, with the assumption that this is the more desirable decision, that is both employees prefer to work in London.

To justify the contrastive decision in such a situation, one has to first show that the preferable decision $d$ had higher probability score for individual $I$ than individual $J$ by enough margin, even when the protected attributes are swapped or made same as shown below:

\begin{align}
           P(\hat Y_{A_i \leftarrow \{a_i, a_j\}\ }(U_i) = d\ |\ X_i = x_i, A_i = a_i)  -
  P(\hat Y_{A_j \leftarrow \{a_i, a_j\}\ }(U_j) = d\ |\ X_j = x_j, A_j = a_j) > \lambda_{ij}  
\end{align}

Besides this, it helps further to justify the decision making process if similar conditions can be shown for $d'$ but with individual $J$ having higher probability than individual $I$, but this is not a necessary condition. Note that combinations of protected attributes are tried by counterfactual logic to verify invariance and hence fairness.

\begin{align}
           P(\hat Y_{A_j \leftarrow \{a_i,a_j\} \ }(U_j) = d'\ |\ X_j = x_j, A_j = a_j)  - 
  P(\hat Y_{A_i \leftarrow \{a_i, a_j\} \ }(U_i) = d'\ |\ X_i = x_i, A_i = a_i) > \lambda_{ij} 
\end{align}

Though fairness of algorithmic decision making is the main objective of the problem at hand, it should ideally be achieved with non-significant loss in algorithmic performance. If the performance error is the difference between the predicted output ($\hat Y$) and expected output ($Y$), then this can be used along with the fairness criteria as a multi-objective pareto front optimisation, with higher priority on fairness in general.

\section{Test Case: Law School Success Revisited}


Though we have formalized contrastive reasoning in the context of algorithmic fairness for the first time in this work, it falls under the purview of causal inference methods, at the heart of which lies interventional and counterfactual logic always. So in order to demonstrate how contrastive logic in fairness builds on the core counterfactual logic, we revisit the same experimental design on law school success as used in [9]. We first point out how the counterfactual fairness evaluation as described in [9] might not be adequate under some conditions. After that we move on to explain how contrastive equations can be utilised to answer further questions using the same framework.

\subsection{Population Level Counterfactual Fairness [9]}

\textbf{Dataset.} The Law School Admission Council dataset [18] has data on 21,790 law students across 163 United States law schools. For each student, it has information like pre-entrance grade-point average (GPA) score, law school entrance examination score (LSAT), post-entrance law school first year grade point average (FYA). The dataset also has some social attributes recorded for students, like race and sex.

\textbf{Problem.} Predict the FYA with sufficient accuracy based on LSAT and GPA while ensuring it is not biased by protected attributes like race and sex.
 
\textbf{Model.} The authors propose 3 levels/types of graphical diagrams to model the problem with some assumptions. Of these, we only consider the highest level (called Level 3 in [9]) in this work, which claims to ensure counterfactual fairness under some strong assumptions. The model is presented in Fig 1  (left diagram). The corresponding structural equations have GPA, LSAT and FYA as functions of race (R), sex (S) [19] and independent error terms as follows:

\begin{align}
\mbox{GPA} &= b_{G} + w_{G}^R R + w_{G}^S S + \epsilon_G, \;\; \epsilon_G \sim p(\epsilon_G) \nonumber \\
\mbox{LSAT} &= b_{L} + w_{L}^R R + w_{L}^S S + \epsilon_L, \;\; \epsilon_L \sim p(\epsilon_L) \nonumber \\
\mbox{FYA} &= b_{F} + w_{F}^R R + w_{F}^S S + \epsilon_F, \;\; \epsilon_F \sim p(\epsilon_F) 
\end{align}

\begin{figure}[th]
  \hspace{-0.3in}
 
   \centerline{\includegraphics[width=1\columnwidth]{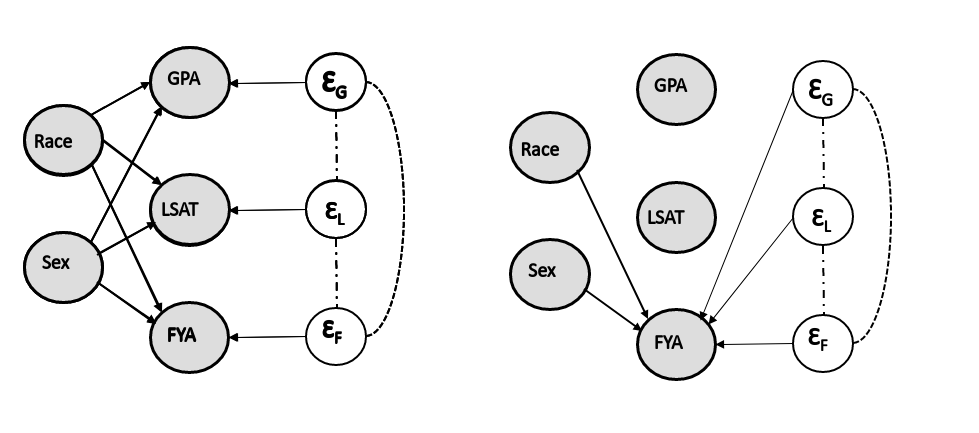}}

  \caption{Left: Causal graphical model of the law school success test case [9]; race and sex are protected variables, $\epsilon_*$ are latent variables, GPA, LSAT and FYA are measured variables. [9] Right: Even when the output variable is directly modeled by latent variables, it might still have non-linear dependance on protected attributes especially by proxy.}
\end{figure}

Since $R$ and $S$ are the protected attributes, the independent error terms in the linear model ($\epsilon_G$, $\epsilon_L$, $\epsilon_F$) are considered to be the latent variables ($U$ from previous sections) the probability distribution of which must be estimated. The terms $b_*$ and $w_*^*$ are the constants and weights respectively of the standard linear models that are fitted to the data. The latent variables, though assumed to be independent of protected variables, may have interactions between them, but this is chosen to be neglected for the model formulation.

Assuming $\mbox{GPA}$ and $\mbox{LSAT}$ be the observed variables, and $\hat {\mbox{GPA}}$ and $\hat {\mbox{LSAT}}$ be their corresponding predictors, then the independent error terms may be estimated as follows. $\epsilon_G$ may be calculated as:

\begin{align}
\hat{\mbox{GPA}} &= b_{G} + w_{G}^R R + w_{G}^S S, \;\;  \epsilon_G = \mbox{GPA} - \hat{\mbox{GPA}} \nonumber \\
\hat{\mbox{LSAT}} &= b_{L} + w_{L}^R R + w_{L}^S S, \;\; \epsilon_L = \mbox{LSAT} - \hat{\mbox{LSAT}}
\end{align}

Thereafter using the estimated values of $\epsilon_G$ and $\epsilon_L$, the predicted value of $\hat {\mbox{FYA}}$ is calculated as:

\begin{align}
\hat{\mbox{FYA}} &= b'_{F} + w_{F}^G \epsilon_G  + w_{F}^L \epsilon_L
\end{align}

The authors [9] claim that the predictor $\hat {\mbox {FYA}}$ thus calculated can be taken to be counterfactually fair since it is only a function of the latent variables that are independent of the protected attributes.

\subsection{Advocating Checking of Individual Level Contrastive Fairness Criteria}

\textbf{Limitations of only using counterfactual fairness.} The counterfactual fairness model makes quite strong assumptions regarding linear relation of protected variables (race and sex) and independence of non-protected variables. This might hold in many cases at a population or even sub-population level of considerable sample size and randomized representation. However, most causal questions of fairness in real life are contrastive and many contrastive questions are asked at individual level. The assumptions of linearity and independence that may hold true at population level might not be valid per individual. For example, consider the attribute:  number of hours of library access after school hours. This might be a latent variable with an effect on LSAT and GPA and over the entire population can be considered to be independent of race and sex. But at an individual level the scenario can be quite different. In fact at an individual level in culturally heterogeneous populations, race might have an effect on whether a particular sex, say female, can avail sufficient mobility after hours and hence this can affect access to library or other resources after hours.

Furthermore, the model itself might need to be different between GPA and LSAT vs. FYA. Eg. the combined effect of race and sex for an individual on after hours mobility and library access changes before and after school. Generally, in school a person lives with their parents and they move out during college. In the latter case guardians have less say and hence mobility will vary a lot between the two cases. Thus the same variable might have a different structural function in first year of law school than earlier in high school. Thus we see that at an individual level there are complex non-linear interactions, interactions by proxy and hence independence is not guaranteed. In fact, in essence it might get reduced to fairness by unawareness.

This is illustrated in the right diagram of fig. 1, where though FYA is modeled using the latent variables in GPA and LSAT, it might have direct dependency to race and sex, specially by proxy. Now these issues can partially be aligned and conceptualised from earlier equations as shown below. FYA is the observed first year grades of the law school students, whereas $\hat {\mbox {FYA}}$ is the predicted value according to equation 17. Now we claim that due to the above discussed effects, the residual $\epsilon_F$ may still be complexly dependent on protected attributes at higher orders:

\begin{equation}
    \epsilon_F = \mbox{FYA} - \hat{\mbox{FYA}}, \;\;  \epsilon_F = b^{''}_{F} + f(R,S) + \epsilon^{''}_F
\end{equation}

Here, $f(R,S)$ is some unknown complex higher order function of the protected variables and $\epsilon^{''}_F$ are the truly independent latent variables. The problem is how to deal with $f(R,S)$, and it is difficult to do that with counterfactual fairness at a population level. We show below that contrastive fairness can be used to mitigate these issues to some extent.

\textbf{Role of contrastive fairness with deep neural representations.} To minimise the effect of higher order interactions of protected attributes on the predictor, the problem needs to be recast as a cost function that might be minimised preferably by a neural network that can represent $f(R,S)$ with sufficient abstraction [20]. This is much easier to do in contrastive case at individual level due to its inherent difference formulation to perform comparison. For individual $I$, the predicted FYA is for simplicity recast below where all the parts independent of race and sex are clumped into $f'_i$.

\begin{equation}
\hat{\mbox{FYA}}_i = f_{i} (R,S) + f^{'}_{i}
\end{equation}

Now when using contrastive logic, we make sure that the decision for individual $i$, is not affected by race and sex at a counterfactual level, that is, we have the cost function:

\begin{align}
\hat{\mbox{FYA}}_i - \hat{\mbox{FYA'}}_i = f_{i} (R,S) -  {f_{i}}_{A_i \leftarrow \{a_{i}^{'}\} \ } (R,S) + f^{'}_{i} - f^{'}_{i}  
= f_{i} (R,S) -  {f_{i}}_{A_i \leftarrow \{a_{i}^{'}\} \ } (R,S)
\end{align}

We use the new formulations to conduct the same experiments on law school data as done in the original counterfactual fairness paper [9]. The results are presented in Table 1. Note the ``Full'' means taking all features available and hence has the highest bias. ``Unaware'' means that the features that are not protected attributes are taken but the fact that they themselves may be biased by proxy by protected attributes is not considered. The code, data and results are directly reproduced from [9] for fair comparison and the new result using contrastive fairness is added.

Here the cost function needs to be minimised with the protected attributes being intervened (using properties of do-calculus from causal inference theory) counterfactually. Since the other terms are independent of race and sex they get cancelled. Representing this by a neural network of sufficient depth to approximate the higher order function $f$ and then minimising the cost function for different individuals as data points can be expected to mitigate the earlier problems to a large extent.

The same logic can be extended when dealing with contrastive predictions between individuals. Given the above discussion, the present authors advocate caution while evaluating population level fairness using counterfactual fairness and suggest that individual level contrastive fairness criteria should also be taken into account with equal importance.

\begin{table}
\centering
\caption{Average accuracy (\%) using logistic regression.}
\begin{tabular}{ccccc} 
\hline
{\bf Full} & {\bf Unaware} & {\bf Counterfactual} & {\bf Contrastive} \\
\hline
0.873 & 0.894 & 0.918 & 0.937 \\
\hline
\end{tabular}
\end{table}

\section{Conclusion}

We adopt contrastive logic from causal inference to solve the question of algorithmic fairness in machine learning. We lay out the mathematical foundations to achieve this with counterfactual logic at its core. Contratstive questions (why this and not that?) have previously been asked in explainable artificial intelligence. But for the first time we propose contrastive criteria (is it fair to take one decision instead of another, differing between two individuals?) in the domain of machine learning. These generic rules can be adopted for various tasks (eg. HR decisions like job recruitment, company layovers, etc). We illustrate the idea with thought experiments and also present initial experimental results on real world  data.

\section{Acknowledgements}
The authors would like to thank Dr. Matt Kusner (University of Oxford and Alan Turing Institute) for his valuable insights. He is one of the authors of the NeurIPS 2017 paper on Counterfactual Fairness [9] and he kindly shared his data and code with us. This also helped the present author for maintaining reproducibility and fair comparison.


\section*{References}

\small






 [1] Miller T.\ (2019) Explanation in Artificial Intelligence: Insights from the Social Sciences. {\it Artificial Intelligence}, 267: 1-38.

[2] Pearl, J.\ (2018) Theoretical Impediments to Machine Learning With Seven Sparks from the Causal Revolution. {\it arXiv:1801.04016}.

[3] Miller, T. (2019) "But why?" Understanding explainable artificial intelligence. {\it ACM Crossroads}, 25(3): 20-25.

[4] DeDeo, S.\ (2016) Wrong side of the tracks: Big Data and Protected Categories. {\it arXiv:1412.4643}.

[5] Angwin, J.; Larson, J.; Mattu, S.; and Kirchner, L. (2016) Machine Bias. {\it ProPublica}.

[6] Silva, R., \ \& Evans, R.\ (2016) Causal inference through a witness protection program. {\it Journal of Machine Learning Research}, 17(56):1–53.

[7] Brennan, T.; Dieterich, W.; and Ehret, B.\ (2009) Evaluating the predictive validity of the compas risk and needs assessment system. {\it Criminal Justice and Behavior}, 36(1):21–40.

[8] Loftus, J. R.; Russell C.; Kusner, M. J.; and Silva R.\ (2018) Causal Reasoning for Algorithmic Fairness. {\it arXiv:1805.05859}.

[9] Kusner M. J.; Loftus, J. R.; Russell C.; and Silva R.\ (2017) Counterfactual Fairness. {\it NeurIPS}, pp.\ 4069-4079.

[10] Lipton, P.\ (1990) Contrastive Explanation. { \it Royal Institute of Philosophy Supplement}. 27:247–266.

[11] Ruben, D.H.\ (1987) Explaining contrastive facts, {\it Analysis}, 47(1):35–37.

[12] Miller, T.\ (2018) Contrastive Explanation: A Structural-Model Approach. {\it arXiv:1811.03163}.

[13] Chouldechova, A. (2017) Fair prediction with disparate impact: a study of bias in recidivism prediction instruments. {\it Big Data}. 2:153–163.

[14] Halpern, J. Y. and Pearl, J.\ (2005) Causes and explanations: A structural-model approach. {\it The British Journal for the Philosophy of Science}. 56(4): 843-911. 

[15] Pearl, J.\ (2009) Causal inference in statistics: An overview. {\it Statistics Surveys}, 3:96–146.

[16] Kusner, M. J.; Russell C.; Loftus, J. R.; and Silva R.\ (2018) Causal Interventions for Fairness. {\it arXiv:1806.02380}.

[17] Bouwel, J. V.  and Weber, E. (2002) Remote causes, bad explanations? {\it Journal for the Theory of Social Behaviour}. 32(4): 437-449.

[18] Wightman, L. F. (1998) Lsac national longitudinal bar passage study. lsac research report series.

[19] Glymour, C., and Glymour, M. R.\ (2014) Commentary: Race and sex are causes. {\it Epidemiology}, 25(4): 488–490.

[20] Zemel, R. S; Wu, Y.; Swersky, K.; Pitassi, T.; and Dwork, C.\ (2013) Learning fair representations. {\it ICML}, 28(3):325–333.

[21] Dhurandhar, A; Chen, P-Y; Luss, R.; Tu, C-C; Ting, P.; Shanmugam, K.; and Das, P. (2018) Explanations based on the Missing: Towards Contrastive Explanations with Pertinent Negatives, \emph{NeurIPS'18}.

\end{document}